\documentclass[preprint,12pt]{elsarticle}
\usepackage{graphicx}
\usepackage{amsmath,amssymb}
\usepackage{xcolor}
\usepackage{float}
\usepackage{multirow}
\usepackage{booktabs}
\usepackage{placeins}
\usepackage{hyperref}
\hypersetup{hypertexnames=false,colorlinks=true,citecolor=blue,linkcolor=blue,urlcolor=blue}
\newcommand{\rev}[1]{#1}
\newcommand{\authorcheck}[1]{#1}
\newcommand{\tablenote}[1]{\par\vspace{3pt}%
  \parbox{\linewidth}{\footnotesize\raggedright #1}}


\journal{Neurocomputing}
\biboptions{sort&compress}

\begin{document}

\begin{frontmatter}

\title{Improved Accuracy Scaling with Problem Complexity in LLM Arithmetic Reasoning Using Executable Symbolic Representations}

\author[aff1]{Yu-Bin Liu}
\author[aff1]{Shi-Ju Ran\corref{cor1}}
\ead{sjran@cnu.edu.cn}
\cortext[cor1]{Corresponding author}

\affiliation[aff1]{organization={Center for Quantum Physics and Intelligent Sciences, Department of Physics, Capital Normal University},
            city={Beijing},
            country={China}}

\begin{abstract}
High accuracy on mathematical benchmarks does not establish whether reasoning
remains reliable as problem complexity increases. We examine how executable
symbolic representations change accuracy scaling in large language models
compared with natural-language reasoning. To address this limitation, we introduce the GSM Arithmetic Benchmark (GSM-AB), a controlled benchmark that independently varies operand digit length and operation count to probe numerical scale and compositional complexity. We find that executable
representations can improve accuracy and slow its decline as complexity
increases. Lean reduces average sensitivity to operand digit length at every tested
model size. By contrast, slower degradation with increasing operation count emerges at the larger model sizes. A Python program-of-thought (Python-PoT) control shows a similar operation-axis pattern, suggesting that this finding is not specific to Lean. These findings show that reasoning representation shapes arithmetic accuracy
scaling and support executable symbolic representations for more reliable
reasoning on complex arithmetic tasks.
\end{abstract}


\end{frontmatter}

\section{Introduction}
\label{sec:introduction}

Aggregate benchmark accuracy does not reveal how large language models (LLMs) respond to increasing mathematical problem complexity. Broader evaluations show that larger, more instructable models do not consistently become more reliable \cite{zhou2024reliability}; mathematical benchmarks also expose degradation under numerical substitutions and logic-preserving increases in difficulty \cite{gsm_symbolic,lpds}. Understanding accuracy scaling with complexity is therefore essential for assessing whether reasoning advances extend to harder problems.

Symbolic methods can complement neural reasoning, as demonstrated by AlphaGeometry in geometry \cite{trinh2024alphageometry}. In arithmetic, Program of Thoughts and Program-Aided Language Models (PAL) improve accuracy by expressing reasoning as executable programs with explicit computational semantics and delegating calculation to an external executor \cite{program_of_thought,pal}. Such executable symbolic representations make numerical computation reliable, but the model must still construct the correct expressions and compose the required steps. Higher accuracy at a given difficulty thus need not imply more favorable scaling as complexity increases.

PAL already provides evidence of robustness to larger numbers and increasing object counts in a symbolic reasoning task \cite{pal}. Other evaluations show that program-of-thought fine-tuning can improve consistency across variants without reducing the accuracy gap under perturbations \cite{pot_abstraction_ceiling}. Building on these findings, we distinguish numerical scale from compositional complexity and examine whether the accuracy advantage of executable symbolic representations over natural-language reasoning is accompanied by slower degradation along each axis, and how this relationship varies with model size. This separation matters because larger operands increase the numerical burden of calculation, whereas additional operations require more arithmetic relations to be composed.

\begin{figure}[t]
    \centering
\includegraphics[width=\textwidth]{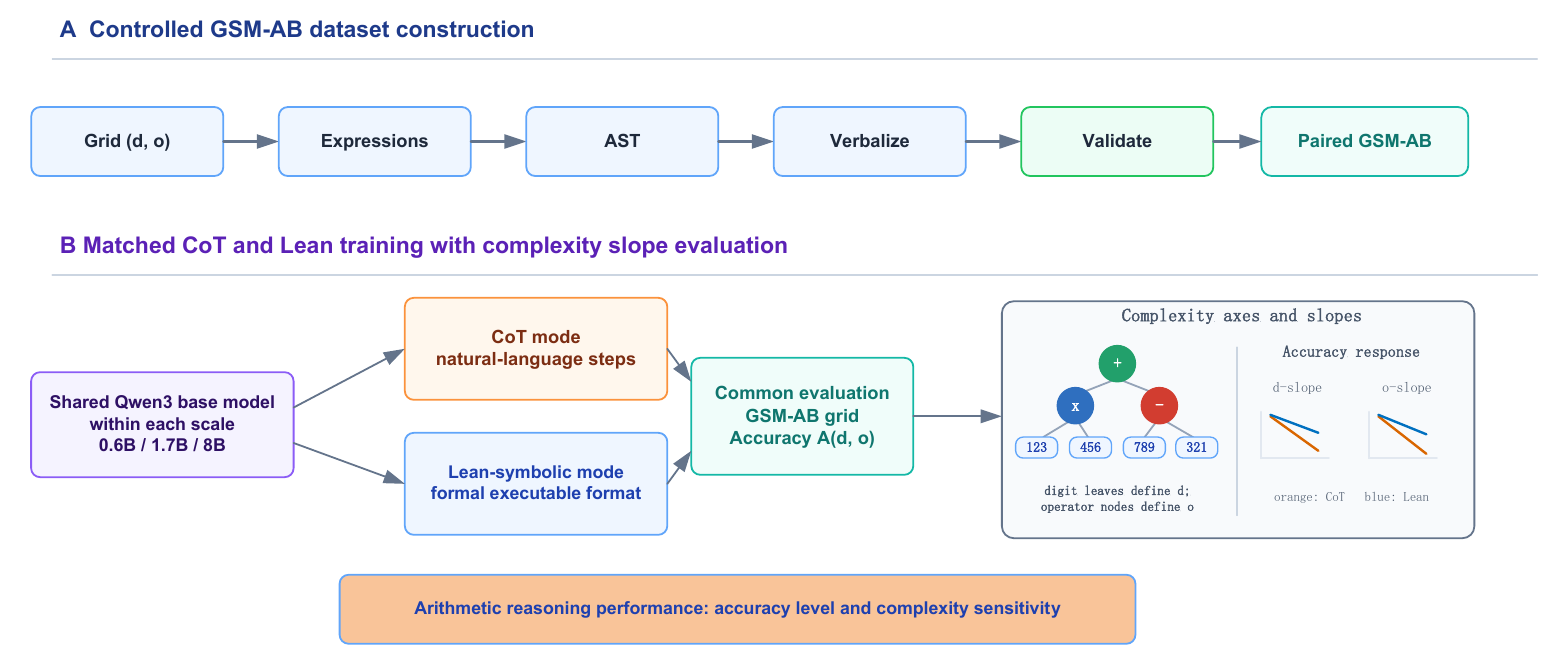}
    \caption{\textbf{Controlled GSM-AB construction and matched chain-of-thought (CoT) and Lean
training with complexity slope evaluation.}
Arithmetic expression trees are generated by independently controlling operand
digit length $d$ and operation count $o$. Each expression is decomposed into an
abstract syntax tree (AST) plan, realized as a GSM8K-style word problem, and
retained only after numerical and solver validation. At each Qwen3 scale,
separate CoT and Lean adapters are trained from the same base model using the
same questions. The loss is computed only on response tokens. The resulting
models are evaluated on the common GSM-AB complexity grid. Accuracy profiles
summarize performance levels, while complexity slopes quantify sensitivity to
operand digit length and operation count. The reported operation slope is
estimated over the common non-floor region. In model-size labels, B denotes billions of parameters.}
    \label{fig:workflow}
\end{figure}

We investigate this question through the GSM Arithmetic Benchmark (GSM-AB), which crosses operand digit length and operation count within a controlled family of balanced expression trees. GSM denotes grade school math, as in the GSM8K word-problem benchmark \cite{gsm8k}. Matching tree structures across digit lengths holds the underlying arithmetic structure fixed, while operation count varies compositional complexity within the template family. Accuracy levels show overall performance, while two fitted slopes show how performance changes as complexity increases. The digit-axis slope measures the change in accuracy as operand digit length increases, and the operation-axis slope measures the change as operation count increases; together, they form the dual-slope framework. Using the same training questions and aligned adaptation settings, we compare natural-language chain-of-thought (CoT) \cite{chain_of_thought} with executable Lean programs across three Qwen3 model sizes. One-shot and external evaluations further assess the scope of the observed differences. Figure~\ref{fig:workflow} summarizes the construction and evaluation protocol; the complete benchmark, training data, and code are publicly available at
\url{https://github.com/Yubinliu-physics/gsm-arithmetic-benchmark}.

Our results show that executable symbolic representations can improve the complexity scaling of arithmetic accuracy under specific conditions. Lean exhibits flatter averaged digit-axis slopes at all three tested model sizes; flatter operation-axis slopes over the common non-floor interval are observed at 1.7B and 8B, where B denotes billions of parameters. The observed scaling improvements vary with the complexity axis and model size. A paired Python--Lean comparison at 8B further finds similar accuracy for both executable representations, substantially above the natural-language baseline. These findings extend the evidence for executable representations from accuracy gains to their dependence on distinct dimensions of arithmetic complexity, with supporting results across two programming languages.

\section{Methods}
\label{sec:method}

\subsection{Controlled arithmetic benchmark}
\label{sec:dataset}
GSM-AB contains 2,000 questions across 20 complexity cells, with 100 questions
for each combination of operand digit length $d\in\{3,4,5,6\}$ and operation
count $o\in\{2,3,4,5,6\}$. Each operation count uses a fixed balanced
binary-tree topology with sampled addition, subtraction, and multiplication.
We first generate and validate six-digit expressions, then resample their
operands to obtain five-, four-, and three-digit counterparts. Operators,
parenthesization, and tree topology remain fixed within each matched group.
Subtractions are constructed to avoid negative intermediate values, yielding
unambiguous integer-valued answers.

Each expression is parsed into an abstract syntax tree (AST) that records
operation dependencies. A language model converts this plan into a GSM8K-style
word problem, preserving the operands and operation order. Questions are
retained after checks of number consistency, answer agreement, and solver
verification. Each record links its complexity metadata, expression, AST,
question, and exact answer (Appendix Fig.~\ref{fig:gsmab_structure}). Wording
is regenerated across digit lengths; generation models and narrative complexity
also vary with operation count. The benchmark therefore controls arithmetic
structure more tightly than linguistic form. Seed pools, generation budgets,
and cross-digit selection are detailed in Appendix~\ref{app:paired_dataset}.

\subsection{Reasoning representations and evaluation}
\label{sec:training}
We compare natural-language CoT with executable symbolic reasoning using
Qwen3 models with 0.6B, 1.7B, and 8B parameters \cite{qwen3}. CoT targets
express intermediate reasoning and the final answer in natural language.
Lean targets encode quantities as typed definitions and arithmetic relations
as expressions evaluated by an external executor. For each size, both branches
start from the same pretrained checkpoint and use low-rank adaptation (LoRA)
\cite{lora} with aligned training settings.

The training pool contains 1,553 paired questions: 1,353 from GSM8K and 200
integer-arithmetic augmentations, including signed subtraction. Each question
has both a natural-language solution and a Lean realization. A seed-42 split
gives 1,398 training and 155 validation questions with identical question
identifiers (IDs) and ordering in both branches. LoRA settings, optimization
schedule, maximum training budget, and the 1,024-token training sequence
limit are shared. Checkpoints are selected independently by the lowest
response-only validation loss, without using GSM-AB; selected steps can
therefore differ between branches. Exact checkpoints and training settings
are listed in Appendix~\ref{app:experimental_details}.

All models are evaluated on the same frozen test items. Main-experiment
decoding is greedy, with generation limits of 256 tokens for CoT and 384 for
Lean. CoT predictions are scored by extracting the final numerical answer;
Lean predictions are compiled and executed with Lean 4.27.0-rc1, then compared
with the reference answer. We also report execution validity to distinguish
execution failures from programs that run but return an incorrect answer.
The Lean demonstration uses \texttt{Int} for integer arithmetic, avoiding
truncating \texttt{Nat} subtraction, and retains \texttt{Rat} for fractions.
This protocol compares complete supervision, representation, and execution
configurations.

A separate Python control compares CoT, Python, and Lean at Qwen3-8B on the
same 500 six-digit questions. All three branches use 1,353 training questions,
matched LoRA settings, and checkpoint 220
(Appendix~\ref{app:python_pot_comparison}). We analyze this control separately
because its training pool and checkpoint selection differ from the main
experiment.

One-shot evaluations assess the representations without parameter updates.
DeepSeek-V4-Flash and Qwen3.5-27B are accessed through application programming
interfaces (APIs) \cite{deepseek_api_models,qwen35} and evaluated on the full
2,000-question grid. Local Qwen3 models use the 500-question $d=6$ subset.
Each mode receives one format-matched demonstration, with internal thinking
disabled and temperature set to zero. CoT and Lean share test questions and
use the answer-scoring procedures described above.

\label{sec:external_generalization}
For external out-of-distribution (OOD) evaluation, we use fixed, ordered
300-example subsets of the SVAMP arithmetic word-problem dataset, DeepMind
Mathematics, and Math401 \cite{svamp,deepmind_math,math401}. Persistent item
IDs are shared across model sizes and reasoning modes. We use the adapters
selected in the main experiment, greedy decoding, and the same 256/384-token
limits. Within each dataset, prompt format, answer extraction, and scoring
remain fixed across model sizes. Generation stops at a complete final-answer
marker or \texttt{\#eval answer}, respectively. These datasets assess accuracy
transfer; their uncontrolled complexity distributions do not support a
separate scaling analysis.

\subsection{Measuring accuracy scaling}
\label{sec:slope}
Let $A_m(d,o)$ denote final-answer accuracy for reasoning mode $m$ at digit
length $d$ and operation count $o$. We characterize two complementary aspects
of complexity sensitivity on the conceptual accuracy response surface. At
fixed $o$, the digit-axis sensitivity is
\begin{equation}
 s^{(d)}_{m,o}=\frac{\partial A_m(d,o)}{\partial d},
\end{equation}
which measures sensitivity to increasing numerical scale. At fixed $d$, the
operation-axis sensitivity is
\begin{equation}
 s^{(o)}_{m,d}=\frac{\partial A_m(d,o)}{\partial o},
\end{equation}
which measures sensitivity to increasing compositional complexity as
operationalized by operation count. These derivatives define the two
axis-specific sensitivity measures independently of a particular benchmark
grid. Because GSM-AB observes $d$ and $o$ only at finite discrete values, we
estimate the corresponding rates of change using ordinary least-squares slopes
fitted to the observed accuracy curves. Cross-model summaries average over the
orthogonal axis before fitting; slice-level slopes hold that axis fixed
(Appendix~\ref{app:slice_slopes}). Digit slopes $s^{(d)}$ use
$d=\{3,4,5,6\}$. Operation slopes $s^{(o,\mathrm{nf})}$ use the common
non-floor interval $o=\{2,3,4\}$, excluding high-operation cells where several
CoT curves flatten near zero accuracy.

Slopes measure accuracy change per added digit or operation; less negative
values indicate slower absolute degradation. We report paired differences
$\Delta d=s^{(d)}_{\mathrm{Lean}}-s^{(d)}_{\mathrm{CoT}}$ and
$\Delta o_{\mathrm{nf}}=s^{(o,\mathrm{nf})}_{\mathrm{Lean}}
-s^{(o,\mathrm{nf})}_{\mathrm{CoT}}$, with positive values favoring Lean.
Throughout this paper, accuracy scaling describes finite-range profiles and
slopes, rather than an assumed power law. Slopes are descriptive point
estimates without confidence intervals and are interpreted alongside full
curves and accuracy levels: unequal baselines and floor or ceiling effects
can influence their magnitudes even within the common non-floor interval.

\section{Results}
\label{sec:results}

\subsection{Accuracy scaling with numerical and compositional complexity}
\begin{figure}[!htbp]
    \centering
    \includegraphics[width=\columnwidth,height=0.65\textheight,keepaspectratio]{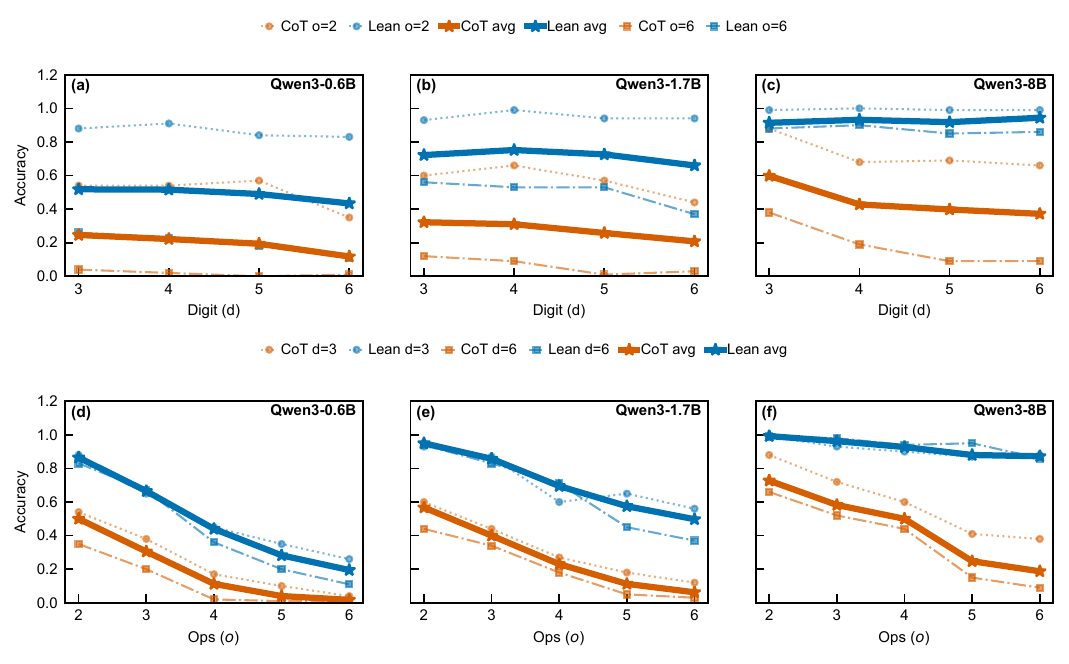}
    \caption{\textbf{Cross-model complexity scaling of CoT and Lean-symbolic
    reasoning.}
    Accuracy is evaluated on the GSM-AB grid for Qwen3-0.6B, Qwen3-1.7B, and
    Qwen3-8B. Top-row panels vary operand digit length $d$ and measure
    numerical-scale sensitivity. Bottom-row panels vary operation count $o$
    and characterize compositional-complexity sensitivity. Thin curves show representative
    fixed-complexity groups, and thick solid curves show averages (avg).
    Ops denotes operation count. Degradation
    slopes summarize how rapidly accuracy changes as arithmetic complexity
    increases. Quantitative slopes are reported in Table~\ref{tab:slope_summary}.
    Operation-axis slopes are fitted only over the non-floor region $o=2,3,4$.}
    \label{fig:cross_model_scaling}
\end{figure}

Executable Lean representations improve mean GSM-AB accuracy over CoT by
29.5, 44.1, and 47.8 percentage points at 0.6B, 1.7B, and 8B, respectively
(Table~\ref{tab:slope_summary}). Their scaling advantage depends on the
complexity axis and model size (Fig.~\ref{fig:cross_model_scaling}).

Along the digit axis, averaged Lean profiles are flatter at all three sizes.
The clearest separation occurs at 8B: CoT accuracy declines with operand
length, whereas Lean accuracy remains approximately stable over $d=3$--$6$.
This pattern also holds within all five operation-count slices at 8B.
At each smaller size, Lean has flatter digit slopes in three of five slices,
with the direction reversing at $o=5$ and $o=6$
(Appendix~\ref{app:slice_slopes}). The numerical-scale benefit is therefore
most consistent at the largest tested size.

\begin{table}[!htbp]
    \caption{Average accuracy differences and degradation slopes across model scales.}
    \label{tab:slope_summary}
    \small
    \renewcommand{\arraystretch}{1.10}
    \begin{tabular*}{\textwidth}{@{\extracolsep{\fill}}lrrrrrrr@{}}
        \toprule
        Model & $\Delta\mathrm{ACC}_{\mathrm{avg}}$ & $s^{(d)}_{\mathrm{CoT}}$ & $s^{(d)}_{\mathrm{Lean}}$ & $\Delta d$
        & $s^{(o,\mathrm{nf})}_{\mathrm{CoT}}$ & $s^{(o,\mathrm{nf})}_{\mathrm{Lean}}$
        & $\Delta o_{\mathrm{nf}}$ \\
        \midrule
        Qwen3-0.6B & $+0.295$ & $-0.041$ & $-0.028$ & $+0.013$ & $-0.194$ & $-0.213$ & $-0.019$ \\
        Qwen3-1.7B & $+0.441$ & $-0.039$ & $-0.021$ & $+0.018$ & $-0.169$ & $-0.128$ & $+0.041$ \\
        Qwen3-8B   & $+0.478$ & $-0.071$ & $+0.008$ & $+0.078$ & $-0.114$ & $-0.033$ & $+0.081$ \\
        \bottomrule
    \end{tabular*}
    \tablenote{\textit{Note:} ACC denotes accuracy; $\Delta\mathrm{ACC}_{\mathrm{avg}}$ is Lean minus CoT
    accuracy averaged over all 20 GSM-AB cells. Slopes are fitted to curves
    averaged over the orthogonal axis: $d=3,4,5,6$ for digit slopes and the
    common non-floor interval $o=2,3,4$ for operation slopes.
    $\Delta d$ and $\Delta o_{\mathrm{nf}}$ are the corresponding Lean-minus-CoT
    slope differences; positive values indicate flatter Lean slopes. Values
    are rounded after calculation. Slopes are descriptive and must be
    interpreted alongside accuracy levels because unequal baselines and
    floor or ceiling effects can affect the comparison.}
\end{table}

Along the operation axis, Lean combines higher accuracy with slower
degradation at 1.7B and 8B. Over $o=2,3,4$, the largest model has a Lean
slope of $-0.033$, compared with $-0.114$ for CoT. At 0.6B, however, Lean
declines slightly faster despite its higher accuracy. Slice-level comparisons
support this model-size dependence: Lean slopes are flatter in all four
digit slices at 8B and in three with one tie at 1.7B, but steeper in three
of four at 0.6B (Appendix~\ref{app:slice_slopes}).

The 0.6B case also separates execution from answer correctness. From $o=2$
to $o=4$, Lean accuracy falls from 0.865 to 0.440, while execution validity
changes only from 0.983 to 0.963
(Appendix~\ref{app:lean_validity_diagnostic}). Most of the accuracy loss
therefore comes from programs that execute but return incorrect answers.
Together, the two axes show that an accuracy gain can accompany improved
numerical scaling without improved compositional scaling.

\subsection{Executable representations beyond Lean}
\label{sec:python_control}
The paired Qwen3-8B control shows a similar operation-axis advantage for
Python and Lean (Table~\ref{tab:python_pot_controlled}). As operation count
increases from two to six, CoT accuracy falls from 0.690 to 0.110, whereas
both executable representations remain at or above 0.860. Over $o=2,3,4$,
the descriptive slopes are $-0.115$ for CoT, $-0.025$ for Python, and
$-0.015$ for Lean. The flatter profiles thus extend beyond a single
executable language.

Pooled accuracy is 0.384 for CoT, 0.950 for Python, and 0.946 for Lean. Python alone answers 16 questions correctly and Lean alone answers 14. The exact two-sided McNemar test finds no statistically significant difference between Python and Lean (\(p=0.8555\)), indicating broadly comparable accuracy under this protocol\cite{fagerland2013mcnemar}. All 500 Python outputs and 499
Lean outputs execute successfully. This control supports a benefit shared
by both executable representations in the tested setting.

\subsection{Scope across prompting settings and tasks}
The advantage also appears in one-shot inference. On the complete GSM-AB
grid, Lean reaches 97.85\% accuracy for DeepSeek-V4-Flash and 97.10\% for
Qwen3.5-27B, exceeding their CoT baselines
(Table~\ref{tab:deepseek_flash_fewshot}). Lean has flatter averaged slopes
along both complexity axes (Appendix~\ref{app:oneshot_slopes}), and its
accuracy advantage grows with operation count in both models. These
profiles extend the main finding to settings without task-specific
fine-tuning.

\begin{table}[!htbp]
    \caption{One-shot accuracy across model families on GSM-AB.}
    \label{tab:deepseek_flash_fewshot}
    \small
    \renewcommand{\arraystretch}{1.10}
    \begin{tabular*}{\textwidth}{@{\extracolsep{\fill}}lccccccc@{}}
        \toprule
        Model & Valid & $d\backslash o$ & 2 & 3 & 4 & 5 & 6 \\
        \midrule
        \multirow{4}{*}{DeepSeek-V4-Flash}
        & \multirow{4}{*}{99.9\%} & 3 & 99/99  & 100/99 & 98/94  & 98/92 & 96/97 \\
        & & 4 & 97/96  & 99/98  & 97/81  & 96/77 & 94/75 \\
        & & 5 & 100/93 & 100/95 & 96/87  & 98/83 & 99/72 \\
        & & 6 & 100/95 & 100/94 & 100/88 & 96/86 & 94/76 \\
        \midrule
        \multirow{4}{*}{Qwen3.5-27B}
        & \multirow{4}{*}{100.0\%} & 3 & 100/99 & 99/97 & 94/89 & 93/85 & 95/87 \\
        & & 4 & 100/95 & 98/92 & 97/91 & 96/80 & 98/82 \\
        & & 5 & 99/84 & 99/71 & 95/56 & 95/42 & 95/31 \\
        & & 6 & 100/73 & 100/58 & 100/51 & 97/27 & 92/21 \\
        \midrule
        Qwen3-0.6B & 72.6\% & 6 & 53/32 & 43/18 & 20/5  & 11/0  & 4/0 \\
        Qwen3-1.7B & 97.0\% & 6 & 89/51 & 87/37 & 68/26 & 51/6  & 42/6 \\
        Qwen3-8B & 99.6\% & 6 & 98/68 & 98/50 & 97/47 & 93/21 & 85/12 \\
        \bottomrule
    \end{tabular*}
    \tablenote{\textit{Note:} Each entry gives the number of correct predictions out of 100 in the form
    Lean/CoT. DeepSeek-V4-Flash and Qwen3.5-27B are evaluated through APIs on
    all 20 cells, whereas the Qwen3 models are evaluated locally at $d=6$. All
    models use the same one-shot prompt format without fine-tuning. All model
    families use the same GSM-AB questions in the cells they share, and CoT and
    Lean are item-paired within every model run. Lean predictions are evaluated by local execution of the
    generated code, whereas CoT predictions are scored from the extracted final
    answer. Valid is the aggregate Lean execution success rate over all cells
    shown for a model: 2,000 outputs for each API model and 500 outputs for each
    local Qwen3 model.}
\end{table}

At $d=6$, all three local Qwen3 models favor Lean in every tested cell.
Performance nevertheless varies strongly with model size. Execution validity
rises from 72.6\% at 0.6B to 99.6\% at 8B, while the larger models retain
substantially higher accuracy as operation count increases. For the larger
local models and both API models, failed execution explains little of the
remaining error.

External accuracy gains are more task dependent. Lean exceeds CoT in seven
of nine dataset--model combinations
(Table~\ref{tab:external_generalization}). The largest gain is on SVAMP at
0.6B ($+0.263$), whereas Lean trails CoT on DeepMind Mathematics at 8B
($-0.050$). On Math401, the methods differ by only one correct answer at
1.7B and 8B. These results broaden the evidence for accuracy gains across
arithmetic tasks while identifying settings in which the advantage shrinks
or reverses.

\begin{table}[!htbp]
    \centering
    \caption{External OOD generalization on arithmetic datasets.}
    \label{tab:external_generalization}
    \small
    \renewcommand{\arraystretch}{1.10}
    \begin{tabular}{@{}lccccc@{}}
        \toprule
        Dataset & Model & CoT & Lean & $\Delta$ & Valid \\
        \midrule
        \multirow{3}{*}{SVAMP}
        & 0.6B & $0.453$ & $0.717$ & $+0.263$ & $0.990$ \\
        & 1.7B & $0.773$ & $0.800$ & $+0.027$ & $0.983$ \\
        & 8B & $0.840$ & $0.880$ & $+0.040$ & $0.990$ \\
  
        \midrule
        \multirow{3}{*}{DeepMind Math}
        & 0.6B & $0.463$ & $0.543$ & $+0.080$ & $0.843$ \\
        & 1.7B & $0.607$ & $0.667$ & $+0.060$ & $0.933$ \\
        & 8B & $0.857$ & $0.807$ & $-0.050$ & $0.893$ \\
        \midrule
        \multirow{3}{*}{Math401}
        & 0.6B & $0.597$ & $0.690$ & $+0.093$ & $0.753$ \\
        & 1.7B & $0.723$ & $0.720$ & $-0.003$ & $0.790$ \\
        & 8B & $0.803$ & $0.807$ & $+0.003$ & $0.837$ \\
        \bottomrule
    \end{tabular}
    \tablenote{\textit{Note:} Each dataset is evaluated on a fixed 300-example subset using Qwen3 models.
    The evaluated datasets are SVAMP~\cite{svamp},
    DeepMind Mathematics~\cite{deepmind_math}, and Math401~\cite{math401}.
    Lean accuracy is computed over all examples after local Lean execution.
    Valid denotes the fraction of generated Lean outputs that execute or verify
    successfully, and $\Delta$ denotes Lean accuracy minus CoT accuracy.}
\end{table}

\section{Conclusion and Discussion}

This work examines how arithmetic accuracy changes as problems become more
complex. We propose a dual-slope analysis. One slope measures the effect of
increasing operand digit length, and the other measures the effect of increasing
operation count. GSM-AB makes these two trends measurable by varying digit
length and operation count separately. Across three Qwen3 model sizes, Lean
achieves higher average accuracy and shows a smaller decline as digit length
increases. For operation count, however, a smaller decline appears only at 1.7B
and 8B. Executable representations therefore improve accuracy, but they do not
improve complexity scaling in every setting.

The difference between the two axes may come from the different demands they
place on the model. Longer operands mainly make numerical calculation harder.
Once the model produces the correct expression, the executor can perform the
calculation. More operations require the model to identify and combine more
arithmetic relations. This remains difficult even when the final calculation
is executable. Python and Lean show similar operation-axis results at 8B,
suggesting that this pattern is not unique to Lean. The one-shot and external
evaluations also show that the observed benefits depend on model size and
dataset.

These conclusions apply to the tested models and arithmetic problems. The
slopes describe the tested model and reasoning representation together, rather
than the model's general mathematical ability. Future work can test other
arithmetic structures, executable languages, and model families.

\section*{Acknowledgements}
We thank Ding Liu for his guidance in preparing the manuscript. We also thank
Pengfei Zhou, Ying Lu, Hao Cui, and Yao Yu for helpful discussions about large
language models.

\section*{Funding}
This work was supported in part by the Strategic Priority Research Program of
the Chinese Academy of Sciences (Grant No.~XDB1270000) and the robotic
AI-Scientist platform of the Chinese Academy of Sciences.

\section*{Data and code availability}
The GSM-AB benchmark (2,000 structure-paired questions), the paired training
data and frozen splits, the complete construction, training, inference, and
verification pipeline, and the frozen experimental results supporting all
tables and figures in this paper are openly available at
\url{https://github.com/Yubinliu-physics/gsm-arithmetic-benchmark}.
\setlength{\bibsep}{0pt plus 0.3ex}
\bibliographystyle{elsarticle-num}
\bibliography{references}

@inproceedings{gsm8k,
  title = {Training Verifiers to Solve Math Word Problems},
  author = {Cobbe, Karl and Kosaraju, Vineet and Bavarian, Mohammad and Chen, Mark and Jun, Heewoo and Kaiser, Lukasz and Plappert, Matthias and Tworek, Jerry and Hilton, Jacob and Nakano, Reiichiro and Hesse, Christopher and Schulman, John},
  booktitle = {arXiv preprint arXiv:2110.14168},
  year = {2021}
}

@inproceedings{gsm_symbolic,
  title = {{GSM-Symbolic}: Understanding the Limitations of Mathematical Reasoning in Large Language Models},
  author = {Mirzadeh, Iman and Alizadeh, Keivan and Shahrokhi, Hooman and Tuzel, Oncel and Bengio, Samy and Farajtabar, Mehrdad},
  booktitle = {International Conference on Learning Representations},
  year = {2025}
}

@inproceedings{svamp,
  title = {Are {NLP} Models Really Able to Solve Simple Math Word Problems?},
  author = {Patel, Arkil and Bhattamishra, Satwik and Goyal, Navin},
  booktitle = {Proceedings of the 2021 Conference of the North American Chapter of the Association for Computational Linguistics: Human Language Technologies},
  pages = {2080--2094},
  year = {2021}
}

@article{lpds,
  title = {{LPDS}: Evaluating {LLM} Robustness Through Logic-Preserving Difficulty Scaling},
  author = {Mondorf, Philipp and Bell, Samuel J. and Dodge, Jesse and Hupkes, Dieuwke},
  journal = {arXiv preprint arXiv:2605.15393},
  year = {2026}
}

@inproceedings{chain_of_thought,
  title = {Chain-of-Thought Prompting Elicits Reasoning in Large Language Models},
  author = {Wei, Jason and Wang, Xuezhi and Schuurmans, Dale and Bosma, Maarten and Ichter, Brian and Xia, Fei and Chi, Ed and Le, Quoc and Zhou, Denny},
  booktitle = {Advances in Neural Information Processing Systems},
  volume = {35},
  pages = {24824--24837},
  year = {2022}
}

@inproceedings{lora,
  title = {{LoRA}: Low-Rank Adaptation of Large Language Models},
  author = {Hu, Edward J. and Shen, Yelong and Wallis, Phillip and Allen-Zhu, Zeyuan and Li, Yuanzhi and Wang, Shean and Wang, Lu and Chen, Weizhu},
  booktitle = {International Conference on Learning Representations},
  year = {2022},
  url = {https://openreview.net/forum?id=nZeVKeeFYf9}
}

@article{qwen3,
  title = {{Qwen3} Technical Report},
  author = {Yang, An and Li, Anfeng and Yang, Baosong and Zhang, Beichen and Hui, Binyuan and others},
  journal = {arXiv preprint arXiv:2505.09388},
  year = {2025}
}

@misc{qwen35,
  title = {{Qwen3.5-27B} Model Card},
  author = {{Qwen Team}},
  year = {2026},
  howpublished = {\url{https://huggingface.co/Qwen/Qwen3.5-27B}},
  note = {Accessed November 3, 2026}
}

@article{deepseek_v32,
  title = {{DeepSeek-V3.2}: Pushing the Frontier of Open Large Language Models},
  author = {{DeepSeek-AI}},
  journal = {arXiv preprint arXiv:2512.02556},
  year = {2025}
}

@misc{deepseek_api_models,
  title = {{DeepSeek API}: Models and Pricing},
  author = {{DeepSeek-AI}},
  year = {2026},
  howpublished = {\url{https://api-docs.deepseek.com/quick_start/pricing}},
  note = {Accessed July 26, 2026}
}

@article{program_of_thought,
  title = {Program of Thoughts Prompting: Disentangling Computation from Reasoning for Numerical Reasoning Tasks},
  author = {Chen, Wenhu and Ma, Xueguang and Wang, Xinyi and Cohen, William W.},
  journal = {Transactions on Machine Learning Research},
  year = {2023}
}

@inproceedings{pal,
  title = {{PAL}: Program-Aided Language Models},
  author = {Gao, Luyu and Madaan, Aman and Zhou, Shuyan and Alon, Uri and Liu, Pengfei and Yang, Yiming and Callan, Jamie and Neubig, Graham},
  booktitle = {Proceedings of the 40th International Conference on Machine Learning},
  series = {Proceedings of Machine Learning Research},
  volume = {202},
  pages = {10764--10799},
  year = {2023}
}

@inproceedings{pot_abstraction_ceiling,
  title = {Program-of-Thought Reveals {LLM} Abstraction Ceilings},
  author = {Zhou, Mike and Bardoliya, Fenil and Gupta, Vivek and Roth, Dan},
  booktitle = {Findings of the Association for Computational Linguistics: {EACL} 2026},
  pages = {4911--4919},
  year = {2026},
  address = {Rabat, Morocco},
  publisher = {Association for Computational Linguistics},
  doi = {10.18653/v1/2026.findings-eacl.257},
  url = {https://aclanthology.org/2026.findings-eacl.257/}
}

@article{deepmind_math,
  title = {Analysing Mathematical Reasoning Abilities of Neural Models},
  author = {Saxton, David and Grefenstette, Edward and Hill, Felix and Kohli, Pushmeet},
  journal = {arXiv preprint arXiv:1904.01557},
  year = {2019}
}

@article{math401,
  title = {How Well Do Large Language Models Perform in Arithmetic Tasks?},
  author = {Yuan, Zheng and Yuan, Hongyi and Tan, Chuanqi and Wang, Wei and Huang, Songfang},
  journal = {arXiv preprint arXiv:2304.02015},
  year = {2023}
}

@article{fagerland2013mcnemar,
  title = {The {McNemar} Test for Binary Matched-Pairs Data: Mid-$p$ and Asymptotic Are Better than Exact Conditional},
  author = {Fagerland, Morten W. and Lydersen, Stian and Laake, Petter},
  journal = {BMC Medical Research Methodology},
  volume = {13},
  number = {1},
  pages = {91},
  year = {2013},
  doi = {10.1186/1471-2288-13-91},
  url = {https://doi.org/10.1186/1471-2288-13-91}
}

@article{zhou2024reliability,
  title = {Larger and More Instructable Language Models Become Less Reliable},
  author = {Zhou, Lexin and Schellaert, Wout and Mart{\'i}nez-Plumed, Fernando and Moros-Daval, Yael and Ferri, C{\`e}sar and Hern{\'a}ndez-Orallo, Jos{\'e}},
  journal = {Nature},
  volume = {634},
  pages = {61--68},
  year = {2024},
  doi = {10.1038/s41586-024-07930-y},
  url = {https://doi.org/10.1038/s41586-024-07930-y}
}

@article{trinh2024alphageometry,
  title = {Solving Olympiad Geometry without Human Demonstrations},
  author = {Trinh, Trieu H. and Wu, Yuhuai and Le, Quoc V. and He, He and Luong, Thang},
  journal = {Nature},
  volume = {625},
  pages = {476--482},
  year = {2024},
  doi = {10.1038/s41586-023-06747-5},
  url = {https://doi.org/10.1038/s41586-023-06747-5}
}

\clearpage
\appendix

\section{Structure-paired construction of GSM-AB}
\label{app:paired_dataset}

\subsection{Design objective and benchmark grid}

\begin{figure}[h]
    \centering
\includegraphics[width=\textwidth]{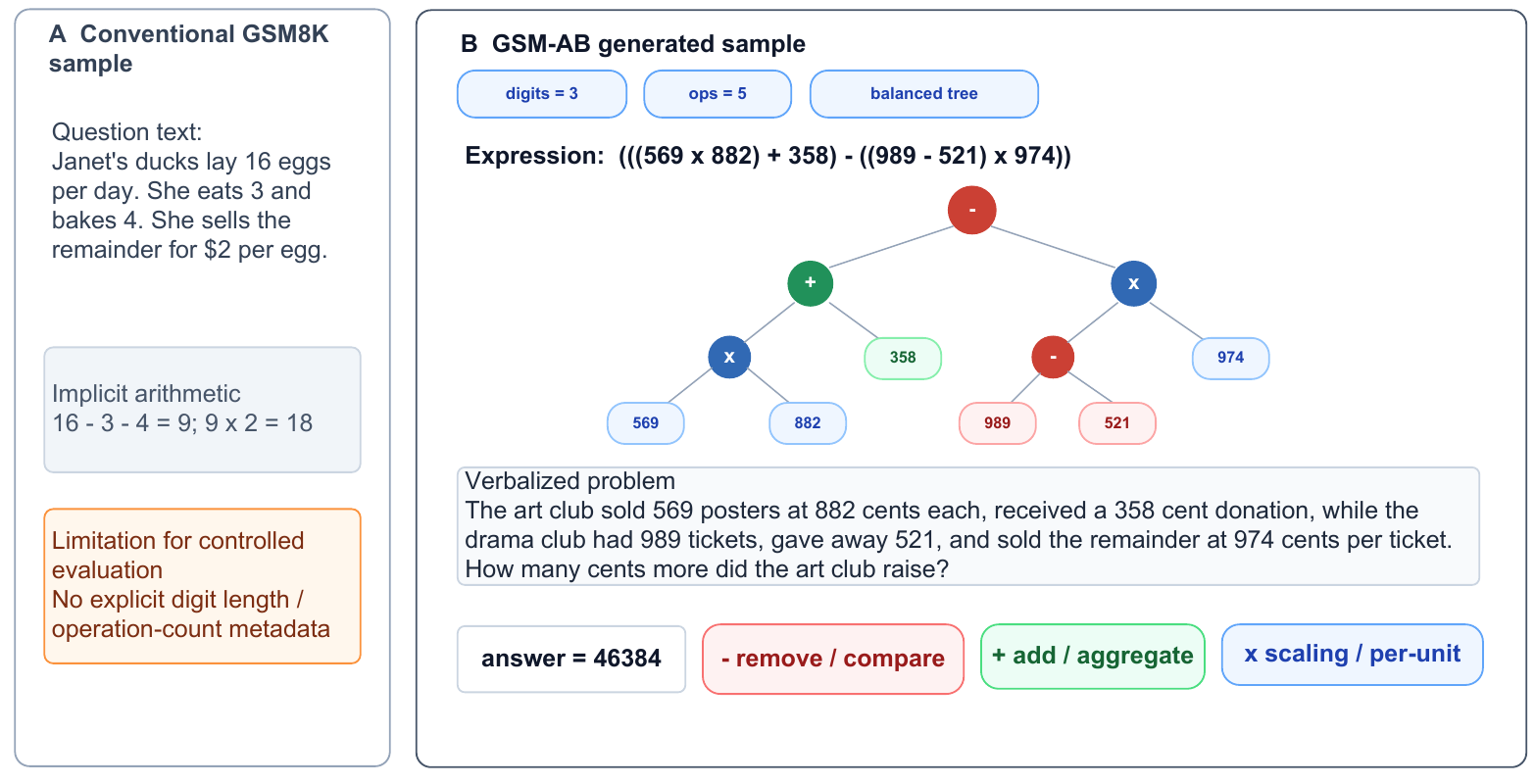}
    \caption{\textbf{Internal structure of a GSM Arithmetic Benchmark
    (GSM-AB) example.}
    The left panel shows a conventional GSM8K problem with a
    natural-language solution and final answer. The right panel shows a
    generated GSM-AB example. Each sample contains explicit complexity
    metadata (operand digit length, operation count, and tree structure), an
    arithmetic expression, its corresponding abstract syntax tree, a
    structure-preserving word problem, and an exact answer. This organization
    makes numerical scale and the associated computational dependency structure
    explicit. Maximum execution depth is recorded and systematically determined
    by the balanced-tree construction. Colors align the symbolic operators with their realizations in
    this generated problem: red marks subtraction through removal and the final
    difference, green marks addition through a received quantity, and blue
    marks multiplication through per-unit rates.}
    \label{fig:gsmab_structure}
\end{figure}

GSM-AB is constructed to vary numerical scale and compositional complexity in
a controlled factorial design. Operand digit length is denoted by
$d\in\{3,4,5,6\}$, and the number of internal arithmetic operations is denoted
by $o\in\{2,3,4,5,6\}$. The final grid contains 100 examples in every $(d,o)$
cell. All examples use recursively constructed balanced binary tree templates.
For a node containing $o$ internal operations, the left subtree contains
$\lfloor o/2\rfloor$ operations and the right subtree contains
$o-1-\lfloor o/2\rfloor$ operations, with the same rule applied recursively.
Thus, every example at a fixed $o$ has the same topology and associated
computational dependency structure. Changing $d$ at fixed $o$ changes only the
numerical magnitudes, not the arithmetic structure being evaluated. We therefore build matched groups
with persistent structure identifiers, rather than generating and filtering
the 20 cells independently. Figure~\ref{fig:gsmab_structure} illustrates the
structure of a representative GSM-AB example with its complexity metadata.

\subsection{Arithmetic expressions and AST plans}

Candidate expressions instantiate these balanced binary tree templates with
addition, subtraction, and multiplication. The leaf nodes are integer operands
whose lengths are controlled by $d$, while the number of internal nodes is
exactly $o$. Operators are sampled with approximately equal probabilities.
Subtraction branches are instantiated so that all intermediate values remain
nonnegative integers. The reference answer is
computed directly from the instantiated expression rather than produced by a
language model.

Each expression is parsed with a deterministic abstract-syntax-tree procedure.
Leaves are assigned execution level zero, and the level of an internal node is
one plus the maximum level of its children. Operations at the same level are
stored as parallel slots, whereas higher-level slots explicitly reference the
completed lower-level subexpressions. This representation preserves
parenthesization and data dependencies and provides the language model with an
ordered arithmetic plan rather than only an unstructured expression string.
Within this fixed template family, $o$ indexes a known dependency structure,
including its maximum execution depth. It therefore varies structural reasoning
demands while holding the topology fixed for a given operation count.
Every seed is assigned a persistent \texttt{structure\_id}. Its source index,
operation count, expression, tree representation, and exact answer are carried
through the subsequent generation and validation stages.

\subsection{Adaptive natural-language realization}

The expression, the AST plan organized by level, and the reference answer are supplied to an
OpenAI-compatible chat-completions model. The model is instructed to output one
GSM8K-style English word problem in a single paragraph, include every source
number exactly once, introduce no additional numerical quantities, preserve
the prescribed operation order, omit the solution and answer, and finish with
a question whose answer equals the reference value.

The realization prompt adapts to both controlled variables. Problems with
$o\leq2$ use short, direct situations. Problems with $o=3$ or $4$ use connected
multi-stage events, while those with $o\geq5$ use explicitly separated branches that are combined only
after their intermediate quantities have been established. Scenario scale is
also adapted to the operands: smaller values permit everyday contexts, whereas
six-digit values use organization-scale settings such as transportation
records, warehouse inventories, website analytics, factory production, or
survey totals. Scenario selection is deterministic for a given input key, so
restarting a run does not silently reshuffle the intended prompt category.

Generation results retain the structural identity and provenance fields and
are sorted by \texttt{structure\_id} and source index after asynchronous API
calls. Empty API content is treated as a failed call and retried. For models
whose default reasoning mode may return an empty content field, internal
thinking is disabled during both generation and verification.

\subsection{Six-digit anchor pool and complexity-adaptive budget}

We construct the paired benchmark from the highest numerical-complexity
condition first. For each $o$, six-digit expressions are realized and verified
before any lower-digit counterpart is derived. We use larger verified reserves
for higher operation counts because complex natural-language realizations fail
more often and because more structures may be lost when the four digit-specific
valid sets are later intersected. Table~\ref{tab:d6_anchor_construction} reports
the realized six-digit anchor construction.

\begin{table}[!htb]
\caption{Construction of the verified six-digit anchor pool.}
\label{tab:d6_anchor_construction}
\small
\renewcommand{\arraystretch}{1.10}
\resizebox{\textwidth}{!}{%
\begin{tabular}{ccccccc}
\toprule
$o$ & Initial & Generated & Passed & Adopted & Final anchor & New model \\
\midrule
2 & 120 & 0   & 0  & 0  & 120 & historical pool \\
3 & 103 & 30  & 18 & 17 & 120 & DeepSeek-V3.2-Instruct \\
4 & 113 & 60  & 46 & 37 & 150 & DeepSeek-V4-Pro \\
5 & 102 & 75  & 48 & 48 & 150 & DeepSeek-V4-Pro \\
6 & 108 & 196 & 95 & 92 & 200 & DeepSeek-V4-Pro \\
\bottomrule
\end{tabular}}
\tablenote{\textit{Note:} ``Initial'' denotes previously validated unique examples. ``Generated'' and
``passed'' refer to additional candidates and solver-verified candidates,
respectively. Only the number needed to reach the pre-specified anchor target
is adopted. For $o=6$, the counts aggregate the main generation round and two
targeted regeneration rounds.}
\end{table}
\FloatBarrier

The target sizes, 120 for $o=2,3$, 150 for $o=4,5$, and 200 for $o=6$, refer
to examples that have already passed validation, not to raw generation calls.
The additional $o=3$ examples are generated and verified with
DeepSeek-V3.2-Instruct \cite{deepseek_v32}. For $o=4,5,6$, both stages use
DeepSeek-V4-Pro \cite{deepseek_api_models} with
internal thinking disabled. \authorcheck{We used the stronger model for the more
complex strata to reduce verification failures and repeated API calls. The
acceptance rules remain unchanged, but model allocation is therefore confounded
with operation count.}

For $o=6$, the main round produces 83 accepted examples from 156 candidates.
Two targeted rounds subsequently recover 6 examples from 25 failed structures
and 6 examples from 15 further candidates. Only 3 examples from the final set
of 6 are needed to reach the target of 200. Targeted regeneration changes only
the natural-language question. The arithmetic expression, exact answer, and
\texttt{structure\_id} remain fixed. This avoids replacing difficult structures
with newly sampled, easier ones.

\subsection{Numerical and solver-based validation}

Validation separates inexpensive deterministic checks from model-based
solution checking. First, the multiset of numerical strings in the generated
question must exactly match that of the source expression, including repeated
values. Questions that introduce, omit, or duplicate a number are rejected
before an API call. We also audit that the expression evaluates to the
stored answer, the number of internal operators equals $o$, every leaf has the
specified digit length, and the question is nonempty.

Candidates passing the numerical checks are then presented to a verifier at
temperature zero. The verifier solves the natural-language problem step by
step and emits a final numerical answer in a constrained format. The extracted
answer is compared with the programmatically computed reference answer using
symbolic normalization. API errors, missing answer fields, answer mismatches,
and numerical-multiset failures are retained with explicit failure reasons.
Only candidates satisfying both stages enter the correct pool. Generation and
verification are separate API calls, although the model capacity is matched to
the operation-count stratum as described above.

The finalized six-digit anchor contains 740 examples: 120 each for $o=2,3$,
150 each for $o=4,5$, and 200 for $o=6$. The release audit confirms 740 unique
\texttt{structure\_id} values, 740 unique expressions, nonempty questions,
exact agreement between expressions and answers, correct operation counts,
six-digit leaf operands, and exact agreement between the numerical multisets in
each question and expression. Existing
validated examples are retained first, and newly verified examples are added
only after deduplication by both \texttt{structure\_id} and normalized
expression.

\subsection{Persistent pairing and joint selection across digit lengths}

Every validated six-digit anchor is used as a mother structure for its
five-, four-, and three-digit variants. For a fixed \texttt{structure\_id}, the
operators, parenthesization, number of internal operations, and ordered
expression-tree topology are held fixed. Only the leaf values are resampled to
the target digit length, subject to the same requirement that intermediate
values remain nonnegative, and the answer is recomputed programmatically. A new
scale-appropriate word problem is then realized and passed through the same
two-stage validation procedure. A failed realization is regenerated for the
same identifier rather than replaced by an independently sampled structure.

After all digit-specific variants have been checked, we compute, separately
for every operation count,

\begin{equation}
\mathcal{C}_{o}=\mathcal{C}_{3,o}\cap\mathcal{C}_{4,o}
\cap\mathcal{C}_{5,o}\cap\mathcal{C}_{6,o},
\end{equation}

where $\mathcal{C}_{d,o}$ is the set of identifiers that pass validation at
digit length $d$ and operation count $o$. We sort the common
\texttt{structure\_id} values lexicographically, select the first 100, and
retain all four digit-specific members of each selected group. This
deterministic rule introduces no model-dependent or run-dependent resampling.
Thus, for each $o$, the frozen benchmark obeys

\begin{equation}
\mathcal{S}_{3,o}=\mathcal{S}_{4,o}=\mathcal{S}_{5,o}=\mathcal{S}_{6,o}.
\end{equation}

The final benchmark contains 20 cells and 2,000 examples. All CoT, Lean, and
evaluations across model scales read the same frozen files. Consequently, comparisons
across digit lengths are made on matched arithmetic structures, and observed
digit-scaling slopes cannot be attributed to structure-distribution drift
caused by independently filtering each $(d,o)$ cell.

\section{Experimental details}
\label{app:experimental_details}

\subsection{Training data and LoRA configuration}

The paired fine-tuning pool contains 1,553 questions: 1,353 GSM8K questions
and 200 augmentations involving integer arithmetic. Every record stores an
answer trace in natural language and a verified Lean4 target, allowing the two
adapters to be trained from identical questions. A fixed seed-42 stratified
split assigns 1,398 questions to training (1,218 original and 180 augmented)
and 155 to validation (135 original and 20 augmented). The same question IDs
and order are used by every model scale and both supervision modes. The prompt
prefix is \texttt{\#\#\# Problem} in both branches, followed by either
\texttt{\#\#\# Answer} or \texttt{\#\#\# Lean4}. A custom causal language-model collator
sets prompt and padding labels to $-100$, so the loss is computed only over the
format-specific response. Checks before training verify the response boundary,
preservation of the mask, and absence of truncation.

Table~\ref{tab:training_configuration} reports the settings shared across model
scales. Both branches use a 1,024-token training limit. Each branch
is trained for at most three epochs and evaluated once per epoch on the fixed
validation split. The selected adapter is the checkpoint with the lowest
response-only validation loss, with earlier checkpoints preferred on an exact
tie. GSM-AB is not consulted during checkpoint selection.

\begin{table}[!hb]
\caption{Fine-tuning and inference configuration for the main experiment.}
\label{tab:training_configuration}
\footnotesize
\setlength{\tabcolsep}{4pt}
\renewcommand{\arraystretch}{1.10}
\resizebox{\textwidth}{!}{%
\begin{tabular}{lcc}
\toprule
Setting & CoT & Lean \\
\midrule
Training pool / train / validation & \multicolumn{2}{c}{1,553 / 1,398 / 155} \\
LoRA rank / alpha / dropout & 8 / 32 / 0.05 & 8 / 32 / 0.05 \\
Target modules & \multicolumn{2}{c}{query/value/gate/down projections} \\
Maximum epochs & 3 & 3 \\
Batch per device / gradient accumulation & 1 / 8 & 1 / 8 \\
Effective batch size & 8 & 8 \\
Learning rate & $2\times10^{-5}$ & $2\times10^{-5}$ \\
Scheduler / warmup ratio & cosine / 0.03 & cosine / 0.03 \\
LoRA bias / task type & \multicolumn{2}{c}{none / causal language modeling} \\
Maximum training length & 1,024 & 1,024 \\
Maximum new tokens in main inference & 256 & 384 \\
Decoding & greedy (\texttt{do\_sample=False}) & greedy (\texttt{do\_sample=False}) \\
Selected steps (0.6B, 1.7B, 8B) & 522, 522, 350 & 522, 522, 522 \\
\bottomrule
\end{tabular}}
\tablenote{\textit{Note:} The CoT and Lean branches use identical questions, LoRA capacity, optimization
schedule, and effective batch size. Different generation limits
reflect the different output formats and are reported explicitly.}
\end{table}
\FloatBarrier

For Qwen3-0.6B, Qwen3-1.7B, and Qwen3-8B, respectively, the selected
CoT checkpoints are steps 522, 522, and 350, with response-only validation
losses of 0.491, 0.405, and 0.302. The corresponding Lean checkpoints are
steps 522, 522, and 522, with validation losses of 0.248, 0.190, and 0.142.
These checkpoints are selected independently for each combination of model and
mode using only the fixed validation split. No GSM-AB result enters the selection
procedure.

\subsection{Model checkpoints and evaluation toolchain}

The main complexity-scaling experiment uses Qwen3-0.6B, Qwen3-1.7B, and the
standard Qwen3-8B checkpoint. CoT and Lean adapters at a given scale always
start from the same checkpoint. The local one-shot experiment also uses these
three standard checkpoints without loading an adapter. The external OOD
generalization evaluation uses the same three standard checkpoints and the
response-only selected CoT and Lean adapters from the main scaling experiment.
Lean outputs are compiled and executed locally. Verification uses Lean
4.27.0-rc1 with a 60-second per-sample timeout. A sample is valid only if
compilation and execution succeed, and is correct only if the resulting value
agrees with the reference answer after numeric normalization.

\subsection{One-shot prompt and decoding protocol}

The local one-shot comparison uses one format-matched demonstration describing
the same tomato-seed arithmetic problem in both modes. The CoT demonstration
ends with a line of the form \texttt{\#\#\#\# <answer>}. The Lean demonstration
defines intermediate values and terminates with exactly one
\texttt{\#eval answer}. The target question is appended using raw text
concatenation. Internal thinking is disabled. Inference uses bfloat16 model
weights, batch size 2, greedy decoding, and a maximum of 700 newly generated
tokens. Generation stops after the first valid final-answer marker in CoT or
after \texttt{\#eval answer} in Lean. Each $d=6$, $o\in\{2,3,4,5,6\}$ cell
contains the same 100 paired questions in the two modes.

DeepSeek-V4-Flash and Qwen3.5-27B are queried through APIs
\cite{deepseek_api_models,qwen35} with the corresponding format-matched
one-shot prompts, thinking disabled, and deterministic decoding. Each API
evaluation covers all 20 GSM-AB cells, using the same 2,000 questions for both
API models. The local Qwen3 evaluations use the same 500 questions from the
$d=6$ cells. \authorcheck{Model-family comparisons are therefore item-matched
in the cells they share, as are the CoT and Lean comparisons within each run.}

\subsection{External OOD dataset subsets and scoring}

Each frozen external OOD evaluation file contains 300 persistent item IDs, and the same ordered IDs are
used for every model scale and both reasoning modes. The subsets are not
resampled during evaluation. SVAMP is scored by normalized numerical equality.
DeepMind Mathematics and Math401 contain direct computations and are scored
with a numerical tolerance of 0.0011, matching the stored evaluation metadata.
Lean validity is determined independently of answer correctness. CoT and Lean
use dataset-matched demonstrations, greedy decoding, and maximum generation
lengths of 256 and 384 tokens, respectively.

\section{Slice-wise complexity slopes}
\label{app:slice_slopes}

\rev{The main analysis fits a slope after averaging over the orthogonal
complexity axis. To check whether those averages are driven by a small number
of conditions, Tables~\ref{tab:digit_slice_slopes} and
\ref{tab:ops_slice_slopes} repeat the fits within each fixed-condition slice.
Digit-axis slopes use $d=3,4,5,6$ at each fixed operation count. Operation-axis
slopes use the same non-floor interval $o=2,3,4$ as the main analysis at each
fixed digit length. In both tables,
$\Delta s=s_{\mathrm{Lean}}-s_{\mathrm{CoT}}$, so a positive value denotes a
flatter Lean slope.}

\begin{table}[!htbp]
\caption{\rev{Digit-axis slopes within fixed-operation slices.}}
\label{tab:digit_slice_slopes}
\centering
\small
\renewcommand{\arraystretch}{1.08}
\begin{tabular*}{0.82\textwidth}{@{\extracolsep{\fill}}ccrrr@{}}
\toprule
Model & Fixed $o$ & $s^{(d)}_{\mathrm{CoT}}$ & $s^{(d)}_{\mathrm{Lean}}$ & $\Delta s^{(d)}$ \\
\midrule
\multirow{5}{*}{0.6B}
& 2 & $-0.054$ & $-0.022$ & $+0.032$ \\
& 3 & $-0.062$ & $+0.004$ & $+0.066$ \\
& 4 & $-0.051$ & $-0.020$ & $+0.031$ \\
& 5 & $-0.028$ & $-0.051$ & $-0.023$ \\
& 6 & $-0.011$ & $-0.050$ & $-0.039$ \\
\midrule
\multirow{5}{*}{1.7B}
& 2 & $-0.057$ & $-0.002$ & $+0.055$ \\
& 3 & $-0.038$ & $-0.011$ & $+0.027$ \\
& 4 & $-0.036$ & $+0.026$ & $+0.062$ \\
& 5 & $-0.031$ & $-0.062$ & $-0.031$ \\
& 6 & $-0.035$ & $-0.057$ & $-0.022$ \\
\midrule
\multirow{5}{*}{8B}
& 2 & $-0.065$ & $-0.001$ & $+0.064$ \\
& 3 & $-0.065$ & $+0.017$ & $+0.082$ \\
& 4 & $-0.050$ & $+0.009$ & $+0.059$ \\
& 5 & $-0.077$ & $+0.024$ & $+0.101$ \\
& 6 & $-0.097$ & $-0.011$ & $+0.086$ \\
\bottomrule
\end{tabular*}
\tablenote{\rev{\textit{Note:} Each slope is an ordinary least-squares fit over $d=3,4,5,6$.}}
\end{table}
\rev{The slice-wise results show both consistency and heterogeneity. For the
digit axis, the positive aggregate difference is reproduced at $o=2,3,4$ for
Qwen3-0.6B and Qwen3-1.7B, but reverses at $o=5,6$; Qwen3-8B has a positive
difference in all five slices. For the operation axis, the negative aggregate
difference at 0.6B is reproduced at three of four digit lengths, with $d=5$ as
the exception. At 1.7B, three slices are positive and the $d=3$ slice is tied;
all four 8B slices are positive. Because ordinary least-squares slopes are
linear in the fitted accuracies, the aggregate slope equals the mean of its
slice slopes. These slice fits therefore expose where the average pattern is
stable and where it changes, but they remain descriptive summaries of the
fixed GSM-AB grid rather than significance tests.}
\begin{table}[!htbp]
\caption{\rev{Non-floor operation-axis slopes within fixed-digit slices.}}
\label{tab:ops_slice_slopes}
\centering
\small
\renewcommand{\arraystretch}{1.08}
\begin{tabular*}{0.82\textwidth}{@{\extracolsep{\fill}}ccrrr@{}}
\toprule
Model & Fixed $d$ & $s^{(o,\mathrm{nf})}_{\mathrm{CoT}}$ & $s^{(o,\mathrm{nf})}_{\mathrm{Lean}}$ & $\Delta s^{(o,\mathrm{nf})}$ \\
\midrule
\multirow{4}{*}{0.6B}
& 3 & $-0.185$ & $-0.215$ & $-0.030$ \\
& 4 & $-0.190$ & $-0.235$ & $-0.045$ \\
& 5 & $-0.235$ & $-0.165$ & $+0.070$ \\
& 6 & $-0.165$ & $-0.235$ & $-0.070$ \\
\midrule
\multirow{4}{*}{1.7B}
& 3 & $-0.165$ & $-0.165$ & $+0.000$ \\
& 4 & $-0.190$ & $-0.110$ & $+0.080$ \\
& 5 & $-0.190$ & $-0.120$ & $+0.070$ \\
& 6 & $-0.130$ & $-0.115$ & $+0.015$ \\
\midrule
\multirow{4}{*}{8B}
& 3 & $-0.140$ & $-0.045$ & $+0.095$ \\
& 4 & $-0.095$ & $-0.025$ & $+0.070$ \\
& 5 & $-0.110$ & $-0.035$ & $+0.075$ \\
& 6 & $-0.110$ & $-0.025$ & $+0.085$ \\
\bottomrule
\end{tabular*}
\tablenote{\rev{\textit{Note:} Each slope is an ordinary least-squares fit over $o=2,3,4$.}}
\end{table}
\FloatBarrier

\section{One-shot slopes for API models}
\label{app:oneshot_slopes}

\rev{Following the main procedure, digit slopes use $d=3,4,5,6$ after averaging
over operation counts, and non-floor operation slopes use $o=2,3,4$ after
averaging over digit lengths. We fit accuracy proportions from each complete API
grid; the estimates are specific to the stated prompt and decoding protocol.}

\begin{table}[h]
\caption{\rev{One-shot dual slopes for API models.}}
\label{tab:oneshot_api_slopes}
\centering
\small
\renewcommand{\arraystretch}{1.10}
\resizebox{\textwidth}{!}{%
\begin{tabular}{lrrrrrr}
\toprule
Model & $s^{(d)}_{\mathrm{CoT}}$ & $s^{(d)}_{\mathrm{Lean}}$ & $\Delta s^{(d)}$
& $s^{(o,\mathrm{nf})}_{\mathrm{CoT}}$ & $s^{(o,\mathrm{nf})}_{\mathrm{Lean}}$
& $\Delta s^{(o,\mathrm{nf})}$ \\
\midrule
DeepSeek-V4-Flash & $-0.0246$ & $+0.0014$ & $+0.0260$ & $-0.0412$ & $-0.0062$ & $+0.0350$ \\
Qwen3.5-27B & $-0.1674$ & $+0.0036$ & $+0.1710$ & $-0.0800$ & $-0.0162$ & $+0.0638$ \\
\bottomrule
\end{tabular}}
\tablenote{\rev{\textit{Note:} Positive $\Delta s=s_{\mathrm{Lean}}-s_{\mathrm{CoT}}$ denotes a flatter Lean slope.}}
\end{table}
\FloatBarrier

\section{Training dynamics on GSM8K}

\begin{figure}[H]
    \centering
    \includegraphics[width=\columnwidth]{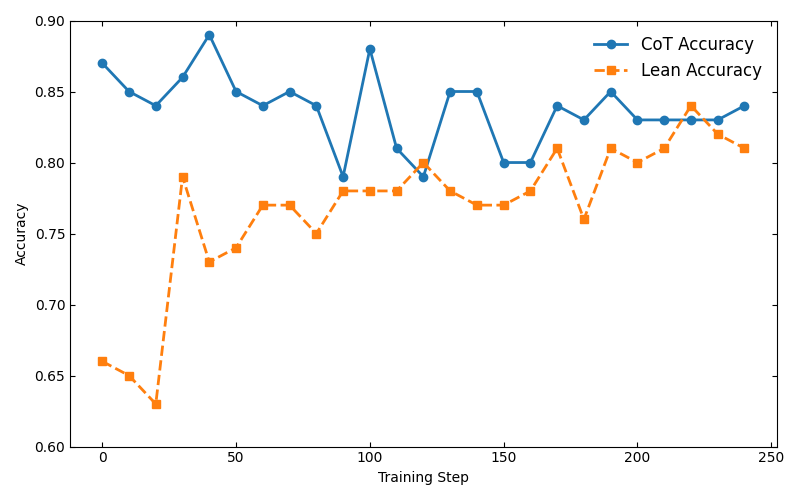}
    \caption{\textbf{Training dynamics of Qwen3-8B on GSM8K.}
    Accuracy is evaluated across LoRA training checkpoints for CoT and
    Lean-symbolic supervision. CoT accuracy fluctuates without a clear monotonic
    gain, whereas Lean-symbolic accuracy increases substantially during
    training.}
    \label{fig:qwen8b_training_dynamics}
\end{figure}

\rev{Figure~\ref{fig:qwen8b_training_dynamics} gives a checkpoint-level view of
Qwen3-8B on original GSM8K. It is included to compare trajectories, not to select
a checkpoint. CoT accuracy is relatively high throughout training and
fluctuates without a clear monotonic trend. The curve therefore shows little
measured improvement over these checkpoints, although accuracy alone cannot
determine which internal capability changed during training.}

\rev{Lean accuracy starts below CoT and rises substantially over training,
eventually approaching the CoT range. Because this evaluation uses original
GSM8K rather than GSM-AB, it complements the main complexity analysis but does
not measure a degradation slope.} \authorcheck{The trajectory is compatible
with gradual acquisition of executable Lean output, but it does not by itself
show that the model internalizes a structured reasoning procedure or that this
procedure causes the GSM-AB gains.}

\section{\rev{Execution validity diagnostic for Qwen3-0.6B}}
\label{app:lean_validity_diagnostic}

\rev{Table~\ref{tab:lean_validity_diagnostic} decomposes the current
Qwen3-0.6B Lean results by operation count. Each row aggregates the four
digit length cells at the corresponding value of $o$ (400 paired questions).
Lean ACC counts a prediction as correct only when the generated program
executes and returns the reference answer. Lean Valid reports only whether the
program executes successfully. It does not establish that the generated
program faithfully represents the word problem. We omit accuracy conditional
on validity because that quantity from the selected subset is not directly comparable
with end-to-end accuracy.}

\begin{table}[!t]
\caption{\rev{Qwen3-0.6B Lean accuracy and execution validity by operation count.}}
\label{tab:lean_validity_diagnostic}
\centering
\small
\renewcommand{\arraystretch}{1.10}
\begin{tabular*}{0.72\textwidth}{@{\extracolsep{\fill}}ccc@{}}
\toprule
$o$ & Lean ACC & Lean Valid \\
\midrule
2 & 0.865 & 0.983 \\
3 & 0.665 & 0.968 \\
4 & 0.440 & 0.963 \\
5 & 0.283 & 0.940 \\
6 & 0.195 & 0.915 \\
\bottomrule
\end{tabular*}
\tablenote{\rev{\textit{Note:} Values aggregate the four digit lengths $d=3,4,5,6$ in the
frozen response-only evaluation. The non-floor slope reported in
Table~\ref{tab:slope_summary} is fitted over $o=2,3,4$.}}
\end{table}

\rev{Across the fitted interval from $o=2$ to $o=4$, accuracy decreases by
0.425 while validity decreases by only 0.020. Non-executable generations
therefore contribute to the decline but cannot account for most of it. The
larger residual consists of executable programs that return an incorrect
answer. End-to-end execution does not reveal whether such a failure arose from
an incorrect translation of the natural-language relations into Lean or from
an incorrect computation encoded within the generated program.}

\section{\rev{Controlled comparison with Python program-of-thought}}
\label{app:python_pot_comparison}

\rev{We add Python program-of-thought (Python-PoT) as a second executable
baseline~\cite{program_of_thought,pal}. The comparison uses the same standard
Qwen3-8B checkpoint and the same 1,353 GSM8K questions for CoT, Python-PoT, and
Lean supervision. The target associated with each question is respectively a
natural-language solution, an execution-verified Python program, or a
Lean program. All three adapters use LoRA rank 8, alpha 32, dropout 0.05, a
learning rate of $2\times10^{-5}$, an effective batch size of 8, two epochs,
16-bit floating-point training, AdamW, and the same target modules. A custom causal language-model collator
preserves the response-only label mask, so prompt tokens do not contribute to
the loss in any branch. We pre-specify checkpoint 220 for all three branches
to provide a matched-training-step control. This 1,353-question control is
separate from the frozen main scaling experiment, in which checkpoints are
selected independently by validation loss.}

\rev{Evaluation uses the same 500 paired GSM-AB questions at $d=6$ and
$o\in\{2,3,4,5,6\}$, with 100 questions per cell. Decoding is greedy and each
format receives one matched demonstration. Python outputs are parsed with a
restricted AST policy and executed in an isolated interpreter. Parsing stops
at the first safe single-variable print statement. Exact adherence to the
requested \texttt{print(answer)} surface form is audited separately and is not
conflated with executability. A prediction is correct only when execution
succeeds and the numerical result matches the reference answer. Lean outputs
are compiled and executed with Lean 4.27.0-rc1 under the same numerical
criterion. For Lean, we retain the same trained checkpoint and change
only the evaluation prompt and matched demonstration: every numerical
definition, including \texttt{answer}, is required to have type \texttt{Int},
and the program ends with \texttt{\#eval answer}. Thus, this is a
prompt-only protocol change rather than a separately retrained model.}

\begin{table}[!t]
\caption{\rev{Controlled comparison of CoT, Python-PoT, and Lean supervision on GSM-AB.}}
\label{tab:python_pot_controlled}
\centering
\small
\renewcommand{\arraystretch}{1.10}
\begin{tabular*}{0.78\textwidth}{@{\extracolsep{\fill}}cccc@{}}
\toprule
$o$ & CoT ACC & Python ACC & Lean ACC \\
\midrule
2 & 0.690 & 1.000 & 0.980 \\
3 & 0.490 & 1.000 & 0.980 \\
4 & 0.460 & 0.950 & 0.950 \\
5 & 0.170 & 0.930 & 0.960 \\
6 & 0.110 & 0.870 & 0.860 \\
\midrule
Overall & 0.384 & 0.950 & 0.946 \\
\bottomrule
\end{tabular*}
\tablenote{\rev{\textit{Note:} All branches use Qwen3-8B, the same 1,353 training
questions, matched LoRA settings, response-only loss, and the same pre-specified
checkpoint 220. This matched-step choice is specific to the controlled
three-branch comparison and is not the checkpoint-selection rule used in the
frozen main scaling experiment. Each non-overall row contains 100 paired
questions at $d=6$. Overall pools all 500 questions. Lean applies the
prompt-only signed-integer protocol described in the text.}}
\end{table}

\rev{Both executable representations maintain substantially higher accuracy
than CoT as operation count increases. CoT accuracy falls from 0.690 at $o=2$ to 0.110
at $o=6$, whereas Python-PoT and Lean remain at or above 0.860 throughout
the tested range. Their pooled accuracies are nearly identical: 0.950 for
Python-PoT and 0.946 for Lean. Python-PoT is higher at $o=2$, $o=3$, and
$o=6$, the methods tie at $o=4$, and Lean is higher at $o=5$. Execution
validity is omitted from Table~\ref{tab:python_pot_controlled} because it is
near ceiling: all 500 Python-PoT outputs executed successfully, and 499 of 500
Lean outputs compiled and executed successfully.}

\rev{Because the two methods answer the same questions, we compare them with a
two-sided exact McNemar test~\cite{fagerland2013mcnemar}, which uses only
paired disagreements rather than treating the two sets of 500 predictions as
independent samples. Python-PoT
alone is correct on 16 questions and Lean alone is correct on 14. Both are
correct on 459 questions and both are wrong on 11. Under the null hypothesis
that the two methods have equal probabilities of winning a discordant pair,
the probability of observing an imbalance at least this extreme is
$p=0.8555$. This $p$-value is not the probability that the methods are
identical. Rather, its large value means that the observed 0.004 accuracy
difference provides no statistically significant evidence of a performance
difference.}

\authorcheck{This controlled experiment supports a benefit of executable
intermediate representations over natural-language CoT on GSM-AB. The small,
non-significant difference between Python and Lean does not support a general accuracy
advantage for either language.}

\end{document}